\documentclass{article}

\usepackage[preprint, nonatbib]{neurips_2024}

\usepackage[backend=biber, style=numeric, maxbibnames=3, minbibnames=3]{biblatex}
\newcommand{\citep}[1]{\parencite{#1}}
\newcommand{\citet}[1]{\textcite{#1}}

\usepackage{graphicx}
\usepackage{tcolorbox}
\usepackage{wrapfig}            

\usepackage[utf8]{inputenc} 
\usepackage[T1]{fontenc}    
\usepackage{hyperref}       
\usepackage{url}            
\usepackage{booktabs}       
\usepackage{amsfonts}       
\usepackage{nicefrac}       
\usepackage{microtype}      
\usepackage{amsmath}

\usepackage{colortbl}
\usepackage{enumerate} 

\usepackage{multirow}
\usepackage{graphicx}
\usepackage{subfig}
\usepackage{censor}        
\usepackage{subcaption}

\usepackage[subtle]{savetrees} 

\definecolor{darkgreen}{rgb}{0.0, 0.6, 0.0}
\newcommand{\col}[1]{\textcolor{darkgreen}{#1}}

\newcommand{\from}{\colon}
\DeclareMathOperator*{\argmax}{arg\,max}

\title{Reverse Image Retrieval Cues Parametric Memory in Multimodal LLMs}

\author{%
  Jialiang Xu\thanks{Equal contribution.} \ $^{1}$ \quad
  \textbf{Michael Moor$^{*1}$\quad}
  \textbf{Jure Leskovec$^{1}$}
  \vspace{0.3cm}\\
  $^{1}$Department of Computer Science, Stanford University \\
  Correspondence to: 
  jure@cs.stanford.edu
}

\StopCensoring 

\begin{document}

\maketitle

\begin{abstract}
    Despite impressive advances in recent multimodal large language models (MLLMs), state-of-the-art models such as from the GPT-4 suite still struggle with knowledge-intensive tasks. To address this, we consider Reverse Image Retrieval~(RIR) augmented generation, a simple yet effective strategy to augment MLLMs with web-scale reverse image search results. RIR robustly improves knowledge-intensive visual question answering~(VQA) of GPT-4V by 37-43\%, GPT-4 Turbo by 25-27\%, and GPT-4o by 18-20\% in terms of open-ended VQA evaluation metrics. To our surprise, we discover that RIR helps the model to better access its \emph{own} world knowledge. Concretely, our experiments suggest that RIR augmentation helps by providing further visual and textual cues without necessarily containing the direct answer to a query.
    In addition, we elucidate cases in which RIR can hurt performance and conduct a human evaluation. Finally, we find that the overall advantage of using RIR makes it difficult for an agent that can choose to use RIR to perform better than an approach where RIR is the default setting.
    
\end{abstract}

\section{Introduction}

\begin{figure}[tbp]
    \includegraphics[width=\linewidth]{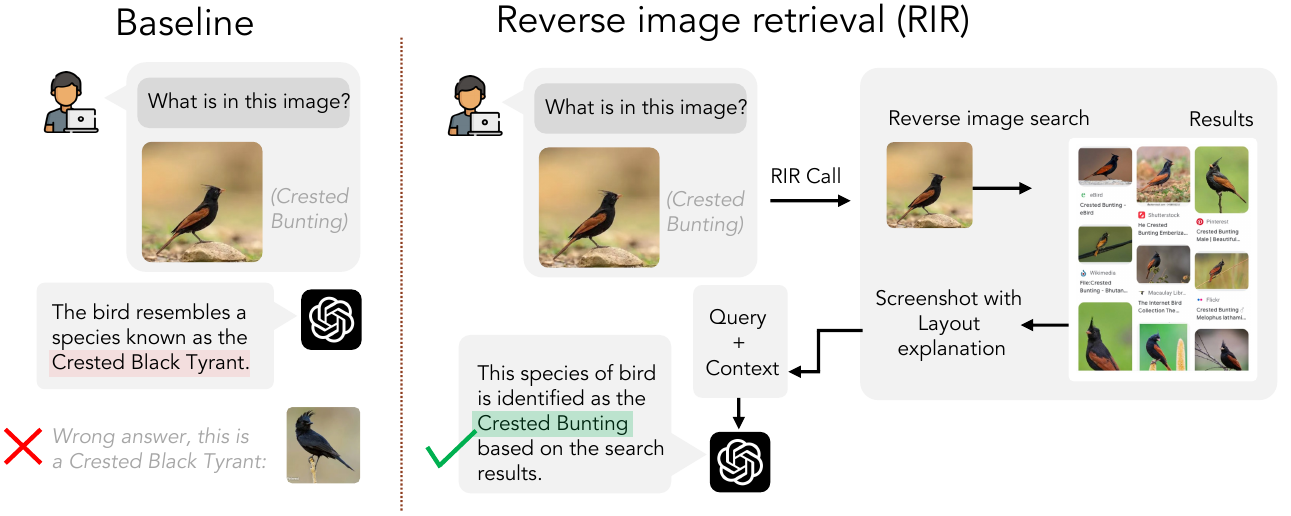}
    \caption{Overview of the reverse image retrieval~(RIR) augmented generation pipeline. In this example, GPT-4V was used as the MLLM backbone. Calling RIR is as easy as running the following line of Python code: \texttt{rir\_api.query\_with\_image(image\_url, query\_text)}. In this basic example leveraging a rare bird species, the correct answer is contained in the search results which may not be the case for knowledge-intensive problems that go beyond the identification of the displayed object.}
    \label{fig:overview}
\end{figure}

\begin{figure}
    \centering
    \includegraphics[width=1.0\linewidth]{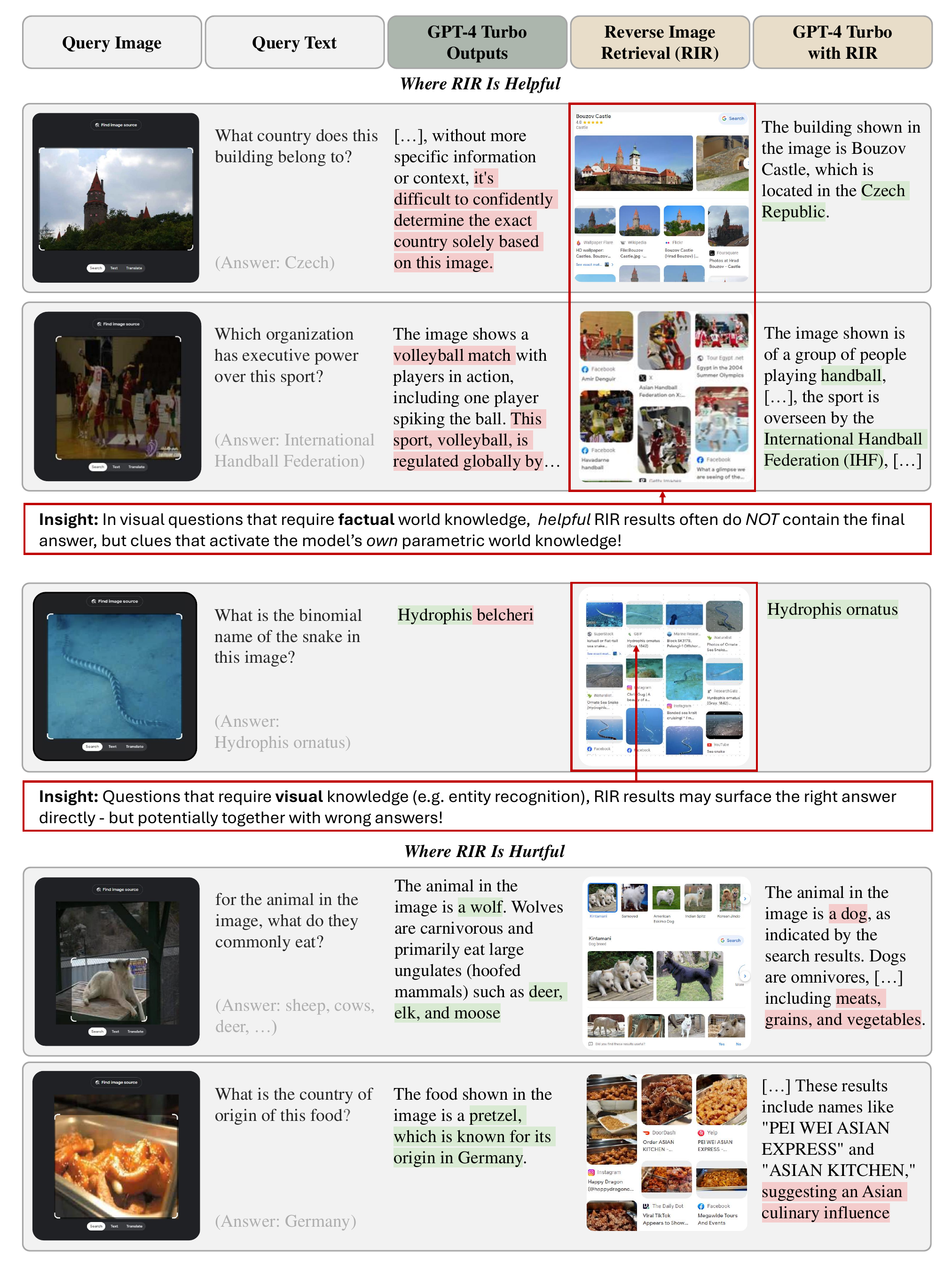} 
    \caption{Selected Output Examples of GPT-4 Turbo before and after being augmented by reverse image retrieval~(RIR). RIR helps identify objects and provide relevant information for unique objects or sites. We also find that in a few cases such as more general and widely-available concepts like certain animals or foods, RIR may be detrimental. Overall, we find that RIR robustly improves the ability of GPT-4-level MLLMs to answer knowledge-intensive visual questions~(Section~\ref{sec:main_results}).
    }
    \label{fig:teaser}
\end{figure}

General-purpose multimodal large language models~(MLLMs), typically vision-language models, have led to significant advances in tasks like visual question answering~\cite{li2023comprehensive, laurenccon2024obelics, liu2024visual, alayrac2022flamingo} and multimodal chatting~\cite{yang2023dawn, wu2023visual, zhu2023minigpt}. However, MLLMs are still struggling with knowledge-intensive tasks, such as answering visual questions that require a large amount of visual knowledge (like recognizing and distinguishing a large number of animal species or medical diagnoses), or that require visual queries to be mapped to textual knowledge (such as stating facts about entities displayed in an image)~\citep{li2023comprehensive, li2023medical, jiang2024evaluating, schmidgall2024agentclinic}.

Many knowledge-intensive multimodal tasks require detailed knowledge about entities that appear in the non-text modality~(e.g. in the image)~\citep{chen-etal-2023-pre-trained}. Aggravatingly, there is a long tail of entities, including objects and concepts that may have little to no support in the multimodal training data distribution used to develop an MLLM, making knowledge-intensive tasks even harder. For example, there are thousands of rare diseases, which were described in only a handful of patients worldwide, thus severely limiting the amount of image samples that could have entered training datasets~\citep{smith2022estimating}. Or to give another example, there are millions of insect species~\cite{mora2011many}, but only a few \emph{dozen} image classes capture insects species within the widely-used ImageNet dataset~\cite{luccioni2023bugs}. While these tasks create the need for a large body of knowledge, existing MLLMs that generate from the limited parametric, multimodal knowledge are underperforming: e.g., they may not recognize a rare bird in an image but would either propose a wrong bird species or reject answering~(see Figure~\ref{fig:overview}), despite the language backbone likely possessing relevant knowledge about the rare bird species of question.

In prior work, a wide range of methods have been proposed to augment LLMs with external memory of text, often referred to as retrieval-augmented generation~(RAG)~\cite{guu2020retrieval, wu2021memorizing, gao2023retrieval}. Augmenting MLLMs with multimodal memory is not yet well understood. While there exist some early efforts in this direction~\cite{sarto2022retrieval, chen2022murag, yasunaga2023retrieval}, they typically leverage relatively small image-text indices, and comparatively small or by now outdated LLM architectures. While recent state-of-the-art LLMs and MLLMs from the GPT-4 suite are equipped with browsing capabilities (i.e., they may access external knowledge sources), it is poorly understood to which degree such models would benefit from explicit augmentation with a multimodal, open-ended external memory. Furthermore, while it has been established that an LLM possesses latent knowledge that may diverge from what the LLM generates~\citep{burns2022discovering}, the knowledge of MLLMs---and their ability to access it---is still poorly understood.

Here, we address these gaps by investigating a simple yet effective strategy: Reverse Image Retrieval~(RIR) augmentation for state-of-the-art MLLMs. Concretely, we build a browser-based API to reverse image search the web. For the sake of simplicity, we capture a screenshot of the multimodal search result comprising multiple result images and captions. The resulting summary image is returned as the search result of this RIR call (for more details, refer to Figure~\ref{fig:overview}) and provided as context to the MLLM. 

In our experiments, we find that RIR robustly and drastically improves knowledge-intensive visual question answering~(VQA) of GPT-4V by 37-43\%, GPT-4 Turbo by 25-27\%, and GPT-4o by 18-20\% in terms of open-ended VQA evaluation metrics. To increase reproducibility, we also explore the benefit of RIR in Idefics-2, a smaller~(8B), open-source MLLM, where we observe more moderate gains of 8-10\%.
We elucidate in which cases RIR can be helpful, and where adding it may be hurtful. 

\begin{figure}
    \centering
    \includegraphics[width=0.9\linewidth]{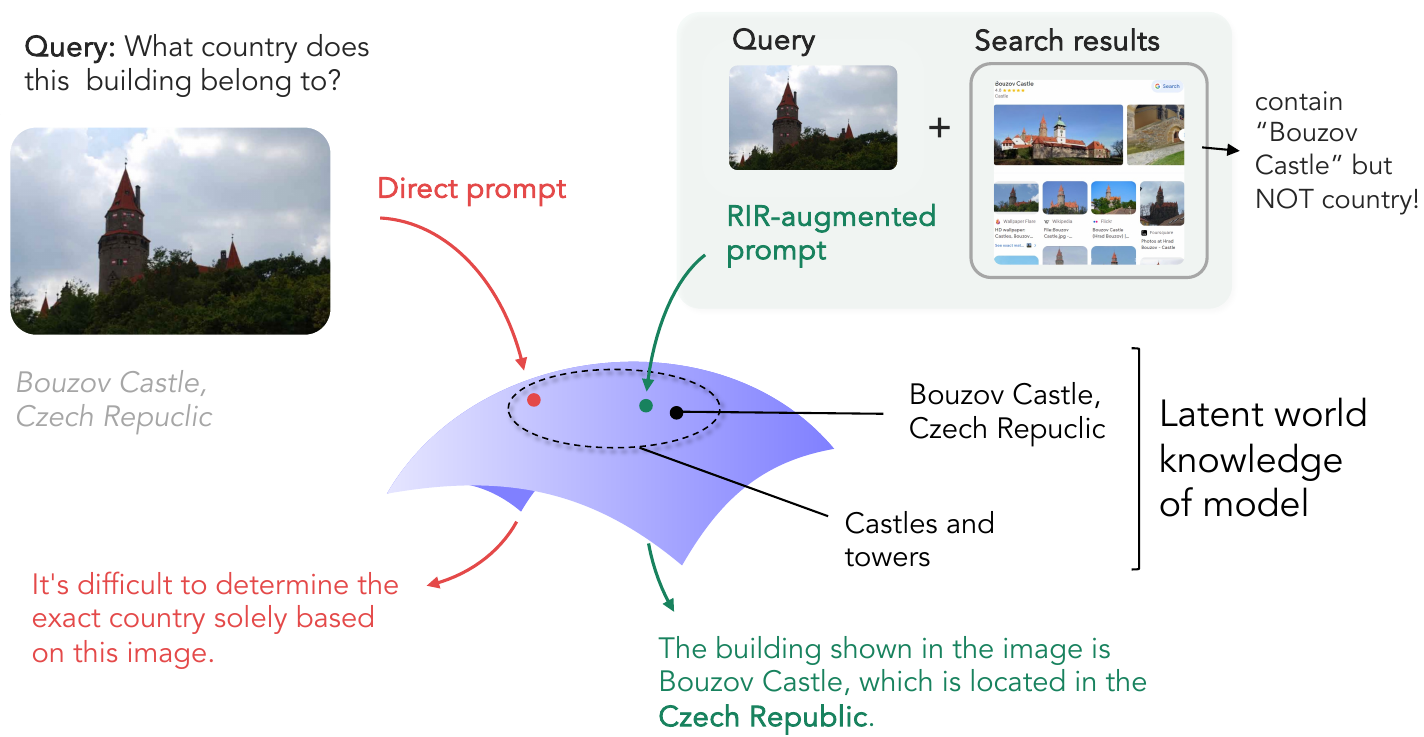}
    \caption{Illustration of reverse image retrieval~(RIR) pipeline. We discover that when prompted with knowledge-intensive visual questions, state-of-the-art MLLMs like GPT-4o can fail to leverage their own world knowledge. Augmenting the query with multimodal RIR results improves the vision-language alignment and allows extraction of highly-specialized text-based knowledge from the model.}
    \label{fig:knowledge}
\end{figure}

To our surprise, we discover that RIR's benefit does not primarily stem from providing the required knowledge to answer knowledge-intensive visual questions, but by improving the alignment of the visual question with the model's own world knowledge~(Figures~\ref{fig:knowledge} and \ref{fig:oracle_entity}). In the INFOSEEK dataset, we observe that the RIR capture does not contain the answer to the factual questions. Instead, RIR offers multimodal cues that assist the MLLM in identifying relevant entities and generating a focused response based on the MLLM's knowledge of those entities. 

We further investigate to which degree gains from RIR correlate with an object's or concept's \emph{lack} of web presence---as a proxy for low support in web-scale multimodal training data distributions. Our findings suggest that RIR helps more with objects and concepts that have less presence on the web. 
Finally, we explore in which scenarios an agent that has the option to call RIR would outperform the approach that defaults to using RIR, and find that for our considered knowledge-intensive VQA benchmarks, the benefit of RIR is so prominent that an agent has to be highly precise in classifying hurtful and helpful cases of using RIR, for which we provide a quantitative guideline. We make all code and data available under \censor{\url{https://github.com/mi92/reverse-image-rag}}.

Overall, this work explores how reverse image retrieval~(RIR) mechanisms can be used to robustly improve the latest state-of-the-art MLLMs. In doing so, we discovered that RIR can improve the alignment of knowledge-intensive visual questions with the parametric world knowledge of MLLMs.

\section{Related works}

\subsection{External memory and LLMs}
Several strategies have been proposed to make external knowledge accessible to large language models~(LLMs). Recent works leveraged \emph{local} external knowledge by making a curated, closed-ended set of external knowledge available in the form of a dedicated external memory, that could be leveraged both during training or during inference~\citep{wu2021memorizing, guu2020retrieval}. By contrast, other works also explored browsing modes of LLMs, i.e., equipping an LLM or LLM-powered system with browsing capabilities by calling a web search API~\citep{openai2024gpt4}. Especially, with the improved quality of the latest state-of-the-art LLMs, placing API calls and processing structured JSON outputs has become feasible to do robustly. 
Finally, an early study introduced an LLM agent that used various tools to handle text and images, but integrating it with a language-only agent made it complex, and it was not made available~\citep{hu2024avis}.

\subsection{Multimodal retrieval augmentation}
Different variants of multimodal retrieval augmentation~(RAG) have been explored in previous works. \citet{sarto2022retrieval} proposed an architecture that for a given query image retrieves the captions of top-k similar images from an external memory of images to then caption the query image. \citet{chen2022murag} and \citet{yasunaga2023retrieval} proposed multimodal RAG approaches that retrieve images and text from multimodal corpora. \citet{caffagni2024wiki} proposed a finetuned LLaVA model that retrieves textual knowledge from Wikipedia. However, these above models were not (yet) released to be included in our study of knowledge-intensive VQA. 
Furthermore, while different strategies for multimodal RAG have been explored, (e.g., image-to-image, image-to-caption, etc.), these prior works leveraged smaller and by now outdated LLMs and relatively ``small'' image corpora featuring millions of images. However, given that there are trillions of images in the web, and likely many billions of images indexed by Google, previous research on MLLMs has not tapped into the full potential of web-scale external multimodal memories that are available today~\cite{photutorial2021photos}.

\section{Reverse Image Retrieval Augmentation for MLLMs}

\subsection{Problem formulation}
Let $v \in V$ be a visual input from the set of all images $V \subset \mathbb{R}^{h \times w}$, for some heights and widths $h,w \in \mathbb{N}$. Let $q \in Q$ denote a question, and $a \in A$ an answer, drawn from the set of all questions $Q$, and answers $A$, respectively. Furthermore, let $j\from V \times Q \times A \times A \to \mathbb{R}$ be a \emph{judge} that for a given visual question $(v,q,a)$ and a generated answer $a^\prime$, provides a score $j(v,q,a,a^\prime) \in \mathbb{R}$.
In this paper, we investigate open-ended, knowledge-intensive visual question answering~(VQA). For this, let us first consider the open-ended VQA problem
\begin{align}
    \argmax_f \ \ \mathbb{E}_{(v, q, a) \sim P(v, q, a)} \left[ j(v,q,a, f(v,q))\right],
    \label{eq:objective}
\end{align}
where $f \from V \times Q \to A$ represents a model (or a compound model system~\citep{compound-ai-blog}) that for a provided visual question $(v,q)$ provides an answer $a'$.
We next consider the knowledge-intensive VQA problem. For sake of generality, we treat the knowledge used in knowledge-intensive VQA in a \emph{procedural} manner, i.e., there is a generic body $K$ that represents knowledge that is queried to retrieve a subset $k \subseteq K$ which is the required knowledge to construct the ground-truth answer $a$. Realizations of $K$ could be unstructured such as a corpus of text, where the subset $k$ would refer to a relevant passage.
$K$ could also be realized with structured knowledge sources as in the form of knowledge graphs, relational databases, or vector indices where $k$ could refer to a subgraph, a query result, or a retrieved vector, respectively.

Now, let $r\from V \times Q \to K$ denote a retrieval function that for a given visual question $(v,q)$ retrieves knowledge $k \subseteq K$. Next, let $g\from V \times Q \times K \to A$ be a map that generates an answer based on the context of the visual question as well as the relevant knowledge. The resulting knowledge-intensive VQA problem can then be stated with Equation~\ref{eq:objective}, where $f$ decomposes to:
\begin{align*}
    f(v,q) = g(v,q, k), \ \ \textrm{with} \ k = r(v,q).
\end{align*}

\subsection{Reverse image retrieval augmentation~(RIR)}
\label{sec:rir-method}

In this paper, we argue that knowledge-intensive VQA is challenging even for the latest state-of-the-art MLLMs such as GPT-4-Turbo or GPT-4o\footnote{for more supporting details, refer to Tables~\ref{tab:infoseek_result} and \ref{tab:snake_result}.} for one of two reasons:
\begin{enumerate}[a)]
    \item the model \emph{lacks} the required knowledge,
    \item the model possesses the knowledge, but does not \emph{leverage} it.
\end{enumerate}

To address the first reason, we propose to enrich knowledge-intensive visual queries for MLLMs with multimodal, up-to-date context retrieved from the open web. While some of the latest LLMs and MLLMs already possess browsing capabilities, the exact routines that were implemented are typically closed-sourced and not publicly known. For instance, the ChatGPT interface suggests that the system is visiting specific web pages. Therefore, we investigate to which degree it is beneficial for MLLMs to augment visual questions with multimodal context sourced from reverse image search results from the web, featuring potentially hundreds of billions of images and captions. We refer to this process as reverse image retrieval~(RIR) augmentation.

For the scope of this study, we implement RIR via a Chromium browser-based API to reverse image search the web by interactively using Google image search. As state-of-the-art MLLMs have become proficient in reading visual text in images, we leverage a straight-forward strategy to process the search results: we capture a screenshot of the multimodal search result comprising multiple result images and captions. The resulting summary image is returned as the search result of this RIR call (for more details, refer to Figure~\ref{fig:overview}). The summary image together with a layout explanation is provided as context to the MLLM~(for details, refer to Section~\ref{supp:prompt-rir}). The choice to leverage a mere screen capture instead of a more fine-grained parsing and multi-hop extraction of information across multiple surfaced web links was motivated by simplicity. However, this choice allowed us to make a surprising discovery about MLLMs, as detailed further in Section~\ref{sec:analysis_own_knowledge}. For a teaser illustration of this finding, refer to Figure~\ref{fig:knowledge}.

\section{Experiments}
In this section, we first provide details about the backbone models (Section~\ref{sec:backbone_models}) and datasets (Section~\ref{sec:datasets}) used to evaluate RIR. Then, we present the experiment results of RIR across all backbone models in Section~\ref{sec:main_results}. In Section~\ref{sec:result_analyses}, we provide analyses and insights about when and why RIR is beneficial. Further details, such as compute resources, prompt details, and additional results are provided in Section~\ref{supp:additional-exp-details} of the appendix.

\subsection{Backbone Models}
\label{sec:backbone_models}
RIR is used to augment the following MLLMs: OpenAI's GPT-4o, GPT-4 Turbo, and GPT-4V~\citep{openai2024gpt4}, and Idefics-2~\citep{laurençon2024matters}. 
OpenAI's GPT-4 models are closed-source and reach impressive performances on multiple visual question-answering datasets, while Idefics-2 is an open-source vision-language model that achieves state-of-the-art performance among open models with less than $10$B parameters~\citep{laurençon2024matters}. 
For the OpenAI GPT-4 models, we use the official API endpoints.\footnote{The specific endpoint versions are: \texttt{gpt-4o-2024-05-13} for GPT-4o, \texttt{gpt-4-turbo-2024-04-09} for GPT-4 Turbo, and \texttt{gpt-4-1106-vision-preview} for GPT-4V.}
For Idefics-2, we use the official model implementation on HuggingFace.\footnote{https://huggingface.co/HuggingFaceM4/idefics2-8b}
We choose these MLLMs to investigate RIR's benefits on open-source and closed-source MLLMs, as well as MLLMs of different levels of vision-language capabilities.

\subsection{Datasets}
\label{sec:datasets}

We evaluate RIR on two knowledge-intense visual question-answering datasets that challenge different aspects of the model's capabilities: \textbf{INFOSEEK} \citep{chen-etal-2023-pre-trained} and \textbf{SnakeCLEF} \citep{picek2023snakeclef}. INFOSEEK comprises fine-grained world knowledge questions spread across eleven categories, enabling comprehensive topic coverage; while SnakeCLEF represents a long-tailed VQA task: the open-ended identification of various (potentially rare) snake species. For INFOSEEK, we use 150 samples per category following \citet{li2023comprehensive}, totaling 1650 INFOSEEK data samples. For SnakeCLEF, we randomly sample 300 images from the validation set. Note that INFOSEEK provides in addition the ground truth Wikidata ID of the entity in each image, which enables our analyses in Section~\ref{sec:result_analyses}.

\paragraph{INFOSEEK.} We employ two metrics to quantify the generation quality of MLLMs: \textit{GPT-as-judge Accuracy} and \textit{Answer-in-prediction Recall}. The metric \textit{GPT-as-judge Accuracy} stands for the percentage of samples where the model-generated response is regarded as correct by GPT-4 Turbo~\citep{zheng2024judging}\footnote{GPT-4 Turbo was selected as we observed fewer judge errors compared to GPT-4o and GPT-4 in small-scale human reviews of the judges.}. Here, GPT-4 Turbo is provided with the original query and the ground truth answer to facilitate contextualization and accurate judgment. The second metric, \textit{Answer-in-prediction Recall}, captures the ratio of data samples whose ground truth answer appears in the model-generated response.

\paragraph{SnakeCLEF.} The ground truth label for each image in SnakeCLEF is the binomial name of the snake in the image, e.g. ``Crotalus viridis'' for the prairie rattlesnake. We use two metrics to capture the model's classification accuracy at different granularities: \textit{Binomial-EM} and \textit{Genus-EM}. 
The \textit{Binomial-EM} metric calculates the percentage of samples where the model-generated response is an exact match of the ground truth binomial name. The second metric \textit{Genus-EM} stands for the ratio of data samples where the genus predicted by the model is an exact match with the ground truth genus. 
In addition, considering the diversity in the formats an MLLM's outputs can take, we calculate two relaxed metrics \textit{Binomial-Recall} and \textit{Genus-Recall}, to capture the percentages of model-generated responses where the ground truth binomial name and the genus appear, respectively.

\subsection{Main Results}
\label{sec:main_results}
We present our experiment results on INFOSEEK in Table~\ref{tab:infoseek_result} and SnakeCLEF in Table~\ref{tab:snake_result}. The 95\% confidence intervals are provided in Section~\ref{sec:bootstrap}.

\begin{table}[h]
\vspace{-0.1cm}
\centering
\caption{
Experiment Results of RIR on INFOSEEK. 
The models and evaluation metrics used are described in Section~\ref{sec:backbone_models} and Section~\ref{sec:datasets}.
We report the average metrics across all eleven categories (``Avg.'') and the relative change in the average metrics (``$\Delta$\%'') displayed in \col{green}.
The best performance in each column is marked in \textbf{bold}. 
}
\label{tab:infoseek_result}
\resizebox{\linewidth}{!}{
\begin{tabular}{lccccccccccccc}
\toprule
\multicolumn{1}{l|}{\textbf{Model}} & \textbf{Avg.} & \multicolumn{1}{c|}{\textbf{$\Delta$\%}} & \textbf{Building} & \textbf{Animal} & \textbf{Plant} & \textbf{Location} & \textbf{Food} & \textbf{OC} & \textbf{Facility} & \textbf{Vehicle} & \textbf{Objects} & \textbf{Sport} & \textbf{Others} \\ \midrule
\multicolumn{14}{c}{\textit{GPT-as-judge Accuracy (\%)}} \\ \midrule
\multicolumn{1}{l|}{Idefics2} & 17.33 & \multicolumn{1}{c|}{-} & 9.33 & 3.33 & 2.67 & 12.67 & 32.00 & 8.00 & 9.33 & 44.67 & 10.00 & 48.00 & 10.67 \\
\multicolumn{1}{l|}{GPT-4V} & 31.33 & \multicolumn{1}{c|}{-} & 19.33 & 30.67 & 8.00 & 24.00 & 41.33 & 18.00 & 26.00 & 63.33 & \textbf{23.33} & 58.00 & 32.67 \\
\multicolumn{1}{l|}{GPT-4-turbo} & 36.61 & \multicolumn{1}{c|}{-} & 29.33 & 37.33 & 12.00 & 35.33 & 47.33 & 23.33 & 40.00 & 60.67 & 18.00 & 60.00 & 39.33 \\
\multicolumn{1}{l|}{GPT-4o} & 39.21 & \multicolumn{1}{c|}{-} & 35.33 & 33.33 & 12.67 & 36.00 & \textbf{54.67} & 22.67 & 40.67 & 65.33 & 21.33 & 63.33 & 46.00 \\
\midrule
\multicolumn{1}{l|}{Idefics2$_{RIR}$} & 18.73 & \multicolumn{1}{c|}{~~\col{8.04~$\uparrow$}} & 19.33 & 5.33 & 2.67 & 17.33 & 35.33 & 5.33 & 13.33 & 44.00 & 4.00 & 47.33 & 12.00 \\
\multicolumn{1}{l|}{GPT-4V$_{RIR}$} & 44.67 & \multicolumn{1}{c|}{\col{42.55~$\uparrow$}} & 54.00 & 35.33 & 20.00 & 46.67 & 46.00 & \textbf{29.33} & 54.67 & 63.33 & 21.33 & 63.33 & \textbf{57.33} \\
\multicolumn{1}{l|}{GPT-4-turbo$_{RIR}$} & 46.42 & \multicolumn{1}{c|}{\col{26.82~$\uparrow$}} & 58.67 & \textbf{38.67} & 20.00 & \textbf{52.67} & 44.00 & 26.67 & 62.00 & 65.33 & 20.00 & 66.00 & 56.67 \\
\multicolumn{1}{l|}{GPT-4o$_{RIR}$} & \textbf{46.91} & \multicolumn{1}{c|}{\col{19.63~$\uparrow$}} & \textbf{59.33} & 33.33 & \textbf{20.67} & 52.00 & 47.33 & 26.00 & \textbf{64.67} & \textbf{68.67} & \textbf{23.33} & \textbf{68.00} & 52.67 \\ \midrule
\multicolumn{14}{c}{\textit{Answer-in-prediction Recall (\%)}} \\ \midrule
\multicolumn{1}{l|}{Idefics2} & 14.18 & \multicolumn{1}{c|}{-} & 6.00 & 4.67 & 4.67 & 10.67 & 20.67 & 8.00 & 9.33 & 26.00 & 8.00 & 48.67 & 9.33 \\
\multicolumn{1}{l|}{GPT-4V} & 29.64 & \multicolumn{1}{c|}{-} & 18.00 & 26.67 & 18.00 & 24.67 & 30.00 & 26.00 & 26.67 & 38.00 & 23.33 & 60.67 & 34.00 \\
\multicolumn{1}{l|}{GPT-4-turbo} & 33.03 & \multicolumn{1}{c|}{-} & 30.00 & 30.00 & 20.67 & 34.67 & 36.00 & 26.67 & 38.67 & 31.33 & 20.67 & 62.00 & 32.67 \\
\multicolumn{1}{l|}{GPT-4o} & 36.00 & \multicolumn{1}{c|}{-} & 33.33 & 30.00 & 28.67 & 36.00 & \textbf{39.33} & 26.67 & 36.67 & 36.67 & 24.00 & 65.33 & 39.33 \\
\midrule
\multicolumn{1}{l|}{Idefics2$_{RIR}$} & 15.64 & \multicolumn{1}{c|}{\col{10.26~$\uparrow$}} & 14.00 & 6.00 & 4.67 & 14.67 & 24.00 & 1.33 & 13.33 & 26.67 & 6.00 & 52.00 & 9.33 \\
\multicolumn{1}{l|}{GPT-4V$_{RIR}$} & 40.73 & \multicolumn{1}{c|}{\col{37.42~$\uparrow$}} & 47.33 & 30.00 & 36.67 & 41.33 & 37.33 & 30.00 & 46.67 & \textbf{40.00} & \textbf{25.33} & 65.33 & 48.00 \\
\multicolumn{1}{l|}{GPT-4-turbo$_{RIR}$} & 41.15 & \multicolumn{1}{c|}{\col{24.59~$\uparrow$}} & 52.67 & \textbf{32.00} & 36.67 & \textbf{48.00} & 34.67 & 27.33 & 48.67 & 39.33 & 20.00 & 67.33 & 46.00 \\
\multicolumn{1}{l|}{GPT-4o$_{RIR}$} & \textbf{42.79} & \multicolumn{1}{c|}{\col{18.86~$\uparrow$}} & \textbf{57.33} & 29.33 & \textbf{38.00} & 45.33 & 37.33 & \textbf{32.67} & \textbf{49.33} & 38.00 & 23.33 & \textbf{70.00} & \textbf{50.00} \\ \bottomrule
\end{tabular}%
}
\end{table}

\paragraph{INFOSEEK.} Augmenting MLLMs with RIR consistently improves the performances of different models on INFOSEEK. As shown in Table~\ref{tab:infoseek_result}, RIR improves state-of-the-art GPT-4o by 19.63\% relative gain on average \textit{GPT-as-judge Accuracy} across all eleven categories. On Idefics-2, RIR provides a more moderate but consistent performance improvement in the majority of categories, resulting in an 8.04\% relative gain. The ablation study provided in Section~\ref{sec:ablation} shows that the benefits of RIR can come from both the relevant texts and images retrieved. We also observe a trend where models that start at a lower INFOSEEK performance benefit from RIR more (comparing the performances of the GPT-4 suite). However, a minimal vision-language capability is required to effectively utilize RIR (comparing Idefics2 with the GPT-4 suite). We hypothesize that the ability to effectively utilize RIR represents an emergent capability, for which we provide more analysis in Section~\ref{sec:analysis_own_knowledge}.

\textbf{Takeaway~1:} 
\textit{RIR robustly improves state-of-the-art MLLMs on knowledge-intensive VQA. Models with lower initial performances benefit more from RIR, provided they can utilize RIR results.}

\begin{table}[h]
\centering
\caption{Experiment Results of RIR on SnakeCLEF.
The models and metrics used are described in Sections~\ref{sec:backbone_models}~and~\ref{sec:datasets} (EM = exact match).
The best performance in each column is marked in \textbf{bold}.}
\label{tab:snake_result}
\resizebox{0.6\textwidth}{!}{%
\begin{tabular}{lcccc}
\toprule
\textbf{Model} & \textbf{\begin{tabular}[c]{@{}c@{}}Binomial\\ EM\end{tabular}} & \textbf{\begin{tabular}[c]{@{}c@{}}Genus\\ EM\end{tabular}} & \textbf{\begin{tabular}[c]{@{}c@{}}Binomial\\ Recall\end{tabular}} & \textbf{\begin{tabular}[c]{@{}c@{}}Genus\\ Recall\end{tabular}} \\ \midrule
Idefics2 & 0.00 & 0.67 & 0.33 & 4.67 \\
GPT4V & 1.67 & 5.00 & 2.00 & 6.33 \\
GPT-4-turbo & 2.33 & 7.33 & 3.00 & 10.00 \\
GPT4o & 5.33 & 20.33 & 5.33 & 20.33 \\
\midrule
Idefics2$_{RIR}$ & 0.00 & 1.33 & 2.00 & 10.67 \\
GPT4V$_{RIR}$ & 11.33 & 24.00 & 11.67 & 24.33 \\
GPT-4-turbo$_{RIR}$ & 11.67 & 24.67 & 12.00 & 25.00 \\
GPT4o$_{RIR}$ & \textbf{12.33} & \textbf{25.67} & \textbf{13.00} & \textbf{26.00} \\ \bottomrule
\end{tabular}%
}
\end{table}

\paragraph{SnakeCLEF.} The SnakeCLEF dataset presents a significant challenge for state-of-the-art MLLMs such as GPT-4o, which provide the exact binomial nomenclature in only 5.33\% of cases. Table~\ref{tab:snake_result} reveals that adding RIR significantly improves MLLMs on SnakeCLEF across all models and almost all metrics. Similar trends to those noted in Takeaway~1 are also observed here. In addition, on the most fine-grained subtask metric, \textit{Binomial-EM}, RIR improves the performance of state-of-the-art GPT-4o by more than $2\times$, from 5.33\% to 12.33\%; and on the most coarse-grained subtask metric, \textit{Genus-Recall}, RIR gives a 27.89\% relative improvement (20.33\% $\rightarrow$ 26.00\%). 

\textbf{Takeaway~2:} \textit{Among tasks that demand knowledge of different granularities, RIR helps more on the task that demands more fine-grained knowledge.}

\subsection{Analysis}
\label{sec:result_analyses}

\subsubsection{RIR helps MLLMs to access their \emph{own} world knowledge}
\label{sec:analysis_own_knowledge}
To determine whether a) MLLMs are limited by their parametric memory, or b) MLLMs struggle with accessing their parametric world knowledge, we conducted the following experiment~(full details in Section~\ref{supp:prompt-oracle} of the appendix). For the INFOSEEK benchmark, we convert visual questions $(v,q)$ to rephrased text-only questions $q'$ that already contain the displayed entity of interest as provided by an oracle~(for details, refer to Section~\ref{supp:prompt-oracle}). 
For example, the question shown in Figure~\ref{fig:knowledge} will be converted into: 
\texttt{Imagine that you are presented with an image of Bouzov Castle. Answer the following question: What country does this building belong to?} 

This oracle-enhanced ablation takes away the challenge of identifying entities and reduces the visual problem to a text-only, factual question about the entities. 
The MLLM performance on this ablation helps distinguish between the two aforementioned hypotheses a) and b) in the following way: if in case a) the model does not possess sufficient knowledge about the entity of interest, its performance on the text-only questions $q'$ will align with its performance on the original visual questions $(v,q)$. Conversely, if in case b) the model does possess knowledge about the entity of interest but cannot access and leverage that knowledge from the visual question $(v,q)$ alone, then we anticipate an improvement in model performance when prompted with the oracle-enhanced question $q'$.

\begin{figure}[h]
    \centering
    \begin{minipage}{0.45\linewidth}
    \includegraphics[width=\textwidth]{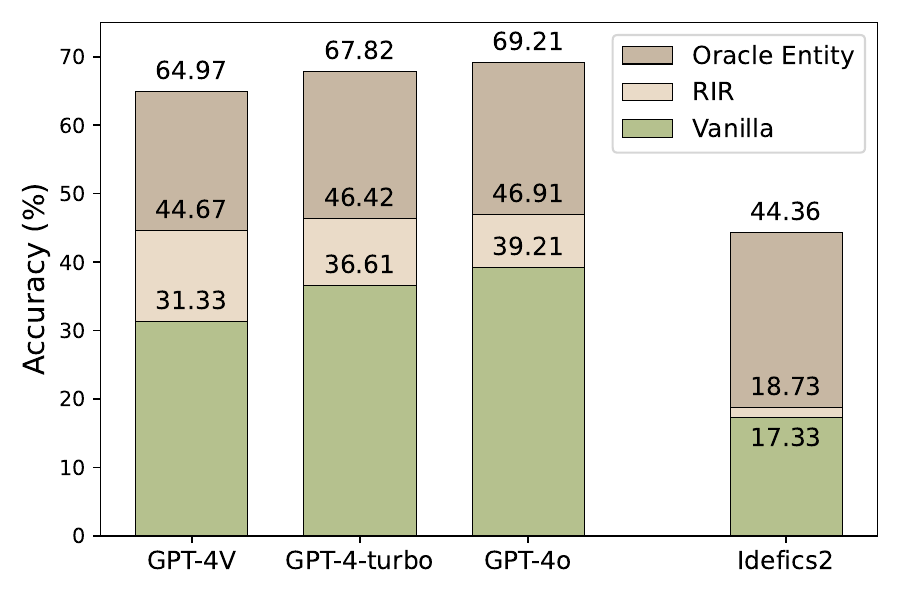}
        \caption{MLLMs provided with text-only questions that contain oracle-provided entities~(brown) show high accuracy on INFOSEEK, showcasing that MLLMs do possess the factual knowledge required, but cannot leverage it in Vanilla VQA prompt. RIR helps close the gap (especially in GPT-4 type models).}
        \label{fig:oracle_entity}
    \end{minipage}
    \hspace{1cm} 
    \begin{minipage}{0.45\linewidth}
    \vspace{-0.8cm}
    \includegraphics[width=1.08\textwidth]{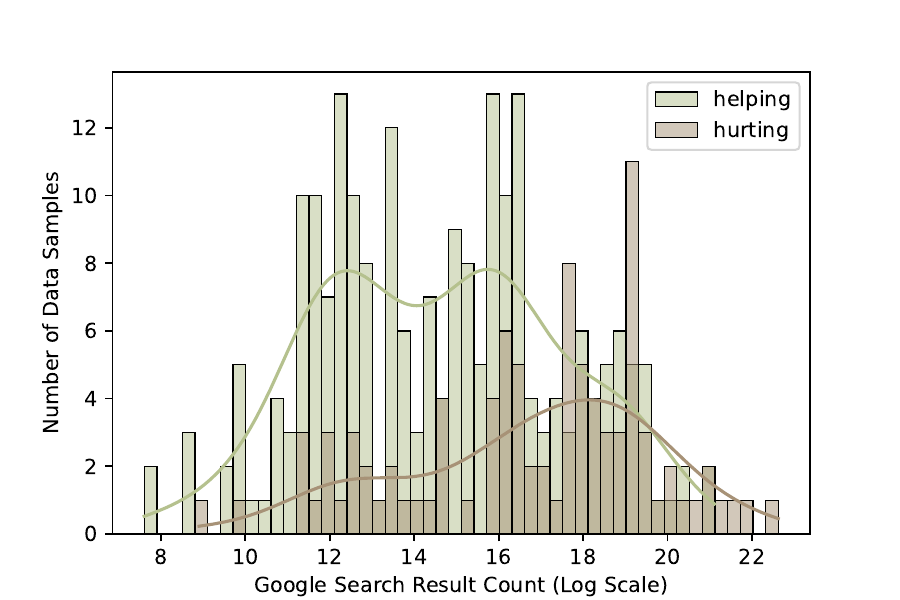}
        \caption{Samples where RIR helps~(green) have fewer Google search results compared to hurting set~(brown), supporting the hypothesis that entities with less online presence are underrepresented in MLLM training datasets, thus benefiting more from RIR's capabilities.
        }
\label{fig:google_search_results}
    \end{minipage}
    
    \vspace{-0.0cm}
\end{figure}

Figure~\ref{fig:oracle_entity} displays the results of this experiment.  
The distinct performance gain of text-only oracle-enhanced variants proves that the MLLM backbones do indeed possess the factual knowledge relevant to answer knowledge-intensive visual questions. However, it seems that when directly prompted with the regular visual question, the model appears unable to properly access and leverage that parametric knowledge. Given that RIR search results do not directly answer the factual questions in INFOSEEK~(see Section~\ref{sec:human-eval}), it suggests that RIR augmentation enhances the MLLM's ability to tap into its \emph{own} parametric knowledge. This is achieved by better aligning the visual question with the extensive world knowledge embedded in the MLLM's language backbone.

When provided with the oracle entity names, all models in Figure~\ref{fig:oracle_entity} show a stark improvement on the metrics. Even the smaller Idefics-2 model that hardly benefits from RIR, possesses more knowledge than the Vanilla VQA prompt would suggest. We hypothesize that Idefics-2 has more problems recognizing entities from the screenshot provided by RIR. To confirm this hypothesis, we conducted an additional evaluation, where we provided the models with the RIR screenshot and prompted them to name the entity in the image. We found that Idefics-2 only correctly recalled $27.76\%$ of the ground truth entity names, while GPT-4o correctly recalled $49.94\%$.

\textbf{Takeaway~3:}
\textit{MLLMs possess more world knowledge than they surface in direct VQA prompts. RIR helps bridge this gap, although its effectiveness depends on the model's ability to accurately interpret the results of RIR.}

\subsubsection{Human evaluation validates GPT-as-judge and RIR stimulating parametric knowledge}\label{sec:human-eval}
Refer to Section~\ref{supp:human-eval} for the full details of our human evaluation of our experiments. In summary, GPT-as-judge Accuracy was aligned with human judgment in $97 \%$ of the reviewed problems regarding the correctness of the model-generated answer. For INFOSEEK, we found that in $99\%$ of cases, RIR results do not contain the final answer, but offer additional context for answering the question.
By contrast, in SnakeCLEF, which is concerned with the open-ended identification of snake species from images, we find that RIR benefits mainly by providing context that partially contains the answer~(in $34\%$ of reviewed cases).

\textbf{Takeaway~4:}
\textit{GPT-as-judge Accuracy reliably aligns with human judgment.}

\subsubsection{RIR helps with long-tail concepts and objects}

We observe that RIR provides the most benefit when the query image is related to entities that are long tail. We filter two sets of data samples from INFOSEEK: the \textit{helping} set and the \textit{hurting} set, where the helping set contains all data samples that the vanilla GPT-4o answered incorrectly and GPT-4o$_{RIR}$ answered correctly, while on the hurting set vanilla GPT-4o answered correctly and GPT-4o$_{RIR}$ answered incorrectly.

As a probing analysis, we use the Google search result count as a surrogate metric for how common an entity is on the Internet. We collect the INFOSEEK-provided ground truth Wikidata ID for each question's entity (described in Section~\ref{sec:datasets}), and fetch its Google search result counts. We show the distribution of Google search result numbers on the helping and hurting sets in Figure~\ref{fig:google_search_results}. 
We found that the distribution of RIR's helping and hurting sets to be significantly different~($p = 0.014$ under Kolmogorov-Smirnov test).
This suggests that RIR indeed helps more on questions about less common entities.

To understand which data types benefit most from RIR, we present the numbers of helping and hurting instances grouped by category in Table~\ref{tab:net_gain_by_category} in the appendix. We observe that for common everyday entities such as food and animals~(like dogs and cats etc.), RIR provides less improvement; and for less common entities such as a unique building or a specific facility, RIR helps MLLMs to answer the question more accurately.

\textbf{Takeaway~5:}
\textit{RIR helps more with long-tail concepts and objects, which might not be sufficiently supported by the multimodal training data of MLLMs.}

\subsubsection{Does a visual search agent outperform RIR?}
\begin{figure}[h!]
    \centering
    \includegraphics[width=0.7\linewidth]{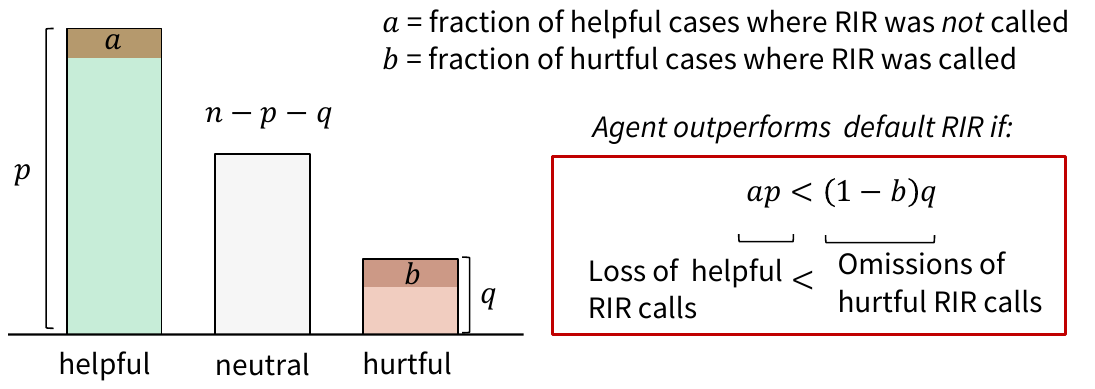}
    \caption{Illustration for when an agent deciding to employ RIR outperforms an approach that defaults to always using RIR. The ideal agent would always call RIR in cases where it is helpful and not call RIR when it is hurtful. $a$ and $b$ denote both classification errors, respectively. For empirical results on agents, refer to Section~\ref{supp:agent}.}
    \label{fig:agent-help}
\end{figure}

Inspired by the recent popularity of web-browsing agents~\citep{deng2024mind2web, liu2023agentbench}, we briefly consider in which cases a basic, MLLM-powered agent that can call RIR is able to improve upon the default RIR-augmented performance~(for experimental results and more details refer to Section~\ref{supp:agent} in the appendix).
For $n$ samples of a given task (e.g., a VQA problem), we distinguish three cases: There are $p$ \emph{helpful} cases where calling RIR is counterfactually flipping a false prediction to a correct prediction. There are $q$ \emph{hurtful} cases, where calling RIR is flipping otherwise correct predictions into wrong ones. Finally, there are $n - q - p$ neutral cases, where the decision to call RIR does not affect the prediction correctness. Also, let $a, b \in [0,1]$ denote the fraction of helpful and hurtful cases that the agent misclassifies, i.e., $a$ is the fraction of helpful cases where RIR was not called, and $b$ is the fraction of hurtful cases where RIR was called.
In Figure~\ref{fig:agent-help}, we illustrate how the decision to call RIR is challenged by the imbalance of $p$ and $q$. In Table~\ref{tab:agent-table} of the appendix, we find empirically that $\frac{p}{q} > \frac{1-b}{a}$, hence an MLLM-based agent that can decide to call RIR in our experiment was sub-par to the approach that always defaults to calling RIR.

\vspace{-0.3cm}
\section{Discussion}
This paper investigates reverse image retrieval~(RIR) augmentation in state-of-the-art multimodal large language models~(MLLMs), specifically in the context of knowledge-intensive visual question answering. We build a simple RIR API to automatically augment MLLMs with multimodal context sourced from the web. In our experiments on two challenging, knowledge-intensive VQA benchmarks---INFOSEEK~(testing factual knowledge about visual objects) and Snake CLEF~(testing visual knowledge directly)---we find that RIR augmentation can drastically improve the performance of state-of-the-art MLLMs.  
 Our original motivation to employ RIR was to augment MLLMs with rich, external knowledge---or to address ``problem a): the lack of knowledge in MLLMs''. However, we discovered that a main benefit of RIR is actually to help align the visual question to the MLLM's \emph{own} textual world knowledge, thereby also helping ``problem b): MLLMs do not fully leverage their parametric knowledge''. Our findings suggest that open-ended entity recognition is relevant for answering knowledge-intensive visual questions and that RIR can serve as an implicit entity recognition step, potentially detecting millions of objects and concepts.

Our study also has limitations. While studying MLLMs from the GPT-4 suite of models is relevant due to their state-of-the-art performance and wide-spread use, the closed-source nature of these models could challenge the reproducibility of our findings (e.g., if the models underlying the API endpoints are changed).

For future work, we will further explore visual search agents. The challenge to overcome will be to increase an agent's accuracy to detect when to use RIR, and when not. We will also explore more fine-grained processing of search results by visiting pages, extract text and images, and investigate to which degree this will add value over the currently implemented screenshot summary.
As Google is concurrently rolling out its new Gemini API, we plan to include any new web-based multimodal RAG solution in our analysis. Finally, it will be an exciting route to investigate whether models that are trained multimodally from scratch (as opposed to fusing separate pre-trained backbones) suffer less from the fragmentation of visual and textual knowledge, that we observed  it in this paper.


\section*{Broader impact}
Augmenting multimodal LLMs~(MLLMs) with multimodal context from reverse image retrieval~(RIR) can potentially have negative societal consequences. For example, if the employed search engine does allow for it, RIR could potentially be abused for surveillance by linking images of persons in public spaces to their web content (such as social media profiles). Adversaries may better geo-locate individual persons based on their social media content which could make them more vulnerable to robberies or scams.
Furthermore, given that RIR can stimulate parametric memory in MLLMs, variants of RIR may be potentially be used in MLLM jailbreaks to reveal sensitive knowledge that is not intended for the user.
As for positive impacts, RIR integrations in MLLM apps could enhance the utility of the MLLM chat interface, e.g. by providing more factually grounded and knowledge-rich answers to visual questions amidst a wide range of application domains and use cases.

\section*{Acknowledgements}
We thank Sam Rawal, Shirley Wu, and Rishabh Ranjan for discussions and for providing feedback on our manuscript. We also gratefully acknowledge the support of
DARPA under Nos. N660011924033 (MCS);
NSF under Nos. OAC-1835598 (CINES), CCF-1918940 (Expeditions), DMS-2327709 (IHBEM);
Stanford Data Applications Initiative,
Wu Tsai Neurosciences Institute,
Stanford Institute for Human-Centered AI,
Chan Zuckerberg Initiative,
Amazon, Genentech, GSK, Hitachi, SAP, and UCB.


\printbibliography

@article{li2023comprehensive,
  title={A comprehensive evaluation of gpt-4v on knowledge-intensive visual question answering},
  author={Li, Yunxin and Wang, Longyue and Hu, Baotian and Chen, Xinyu and Zhong, Wanqi and Lyu, Chenyang and Zhang, Min},
  journal={arXiv preprint arXiv:2311.07536},
  year={2023}
}

@article{wu2023visual,
  title={Visual chatgpt: Talking, drawing and editing with visual foundation models},
  author={Wu, Chenfei and Yin, Shengming and Qi, Weizhen and Wang, Xiaodong and Tang, Zecheng and Duan, Nan},
  journal={arXiv preprint arXiv:2303.04671},
  year={2023}
}

@article{zhu2023minigpt,
  title={Minigpt-4: Enhancing vision-language understanding with advanced large language models},
  author={Zhu, Deyao and Chen, Jun and Shen, Xiaoqian and Li, Xiang and Elhoseiny, Mohamed},
  journal={arXiv preprint arXiv:2304.10592},
  year={2023}
}

@article{laurenccon2024obelics,
  title={Obelics: An open web-scale filtered dataset of interleaved image-text documents},
  author={Lauren{\c{c}}on, Hugo and Saulnier, Lucile and Tronchon, L{\'e}o and Bekman, Stas and Singh, Amanpreet and Lozhkov, Anton and Wang, Thomas and Karamcheti, Siddharth and Rush, Alexander and Kiela, Douwe and others},
  journal={Advances in Neural Information Processing Systems},
  volume={36},
  year={2024}
}

@article{mora2011many,
  title={How many species are there on Earth and in the ocean?},
  author={Mora, Camilo and Tittensor, Derek P and Adl, Sina and Simpson, Alastair GB and Worm, Boris},
  journal={PLoS biology},
  volume={9},
  number={8},
  pages={e1001127},
  year={2011},
  publisher={Public Library of Science}
}

@inproceedings{luccioni2023bugs,
  title={Bugs in the data: How ImageNet misrepresents biodiversity},
  author={Luccioni, Alexandra Sasha and Rolnick, David},
  booktitle={Proceedings of the AAAI Conference on Artificial Intelligence},
  volume={37},
  number={12},
  pages={14382--14390},
  year={2023}
}

@inproceedings{chen-etal-2023-pre-trained,
    title = "Can Pre-trained Vision and Language Models Answer Visual Information-Seeking Questions?",
    author = "Chen, Yang  and
      Hu, Hexiang  and
      Luan, Yi  and
      Sun, Haitian  and
      Changpinyo, Soravit  and
      Ritter, Alan  and
      Chang, Ming-Wei",
    editor = "Bouamor, Houda  and
      Pino, Juan  and
      Bali, Kalika",
    booktitle = "Proceedings of the 2023 Conference on Empirical Methods in Natural Language Processing",
    month = dec,
    year = "2023",
    address = "Singapore",
    publisher = "Association for Computational Linguistics",
    url = "https://aclanthology.org/2023.emnlp-main.925",
    doi = "10.18653/v1/2023.emnlp-main.925",
    pages = "14948--14968",
}

@inproceedings{wu2021memorizing,
  title={Memorizing Transformers},
  author={Wu, Yuhuai and Rabe, Markus Norman and Hutchins, DeLesley and Szegedy, Christian},
  booktitle={International Conference on Learning Representations},
  year={2021}
}

@article{gao2023retrieval,
  title={Retrieval-augmented generation for large language models: A survey},
  author={Gao, Yunfan and Xiong, Yun and Gao, Xinyu and Jia, Kangxiang and Pan, Jinliu and Bi, Yuxi and Dai, Yi and Sun, Jiawei and Wang, Haofen},
  journal={arXiv preprint arXiv:2312.10997},
  year={2023}
}

@misc{openai2024gpt4,
      title={GPT-4 Technical Report}, 
      author={OpenAI and Josh Achiam and Steven Adler and Sandhini Agarwal and Lama Ahmad and Ilge Akkaya and Florencia Leoni Aleman and Diogo Almeida and Janko Altenschmidt and Sam Altman and Shyamal Anadkat and Red Avila and Igor Babuschkin and Suchir Balaji and Valerie Balcom and Paul Baltescu and Haiming Bao and Mohammad Bavarian and Jeff Belgum and Irwan Bello and Jake Berdine and Gabriel Bernadett-Shapiro and Christopher Berner and Lenny Bogdonoff and Oleg Boiko and Madelaine Boyd and Anna-Luisa Brakman and Greg Brockman and Tim Brooks and Miles Brundage and Kevin Button and Trevor Cai and Rosie Campbell and Andrew Cann and Brittany Carey and Chelsea Carlson and Rory Carmichael and Brooke Chan and Che Chang and Fotis Chantzis and Derek Chen and Sully Chen and Ruby Chen and Jason Chen and Mark Chen and Ben Chess and Chester Cho and Casey Chu and Hyung Won Chung and Dave Cummings and Jeremiah Currier and Yunxing Dai and Cory Decareaux and Thomas Degry and Noah Deutsch and Damien Deville and Arka Dhar and David Dohan and Steve Dowling and Sheila Dunning and Adrien Ecoffet and Atty Eleti and Tyna Eloundou and David Farhi and Liam Fedus and Niko Felix and Simón Posada Fishman and Juston Forte and Isabella Fulford and Leo Gao and Elie Georges and Christian Gibson and Vik Goel and Tarun Gogineni and Gabriel Goh and Rapha Gontijo-Lopes and Jonathan Gordon and Morgan Grafstein and Scott Gray and Ryan Greene and Joshua Gross and Shixiang Shane Gu and Yufei Guo and Chris Hallacy and Jesse Han and Jeff Harris and Yuchen He and Mike Heaton and Johannes Heidecke and Chris Hesse and Alan Hickey and Wade Hickey and Peter Hoeschele and Brandon Houghton and Kenny Hsu and Shengli Hu and Xin Hu and Joost Huizinga and Shantanu Jain and Shawn Jain and Joanne Jang and Angela Jiang and Roger Jiang and Haozhun Jin and Denny Jin and Shino Jomoto and Billie Jonn and Heewoo Jun and Tomer Kaftan and Łukasz Kaiser and Ali Kamali and Ingmar Kanitscheider and Nitish Shirish Keskar and Tabarak Khan and Logan Kilpatrick and Jong Wook Kim and Christina Kim and Yongjik Kim and Jan Hendrik Kirchner and Jamie Kiros and Matt Knight and Daniel Kokotajlo and Łukasz Kondraciuk and Andrew Kondrich and Aris Konstantinidis and Kyle Kosic and Gretchen Krueger and Vishal Kuo and Michael Lampe and Ikai Lan and Teddy Lee and Jan Leike and Jade Leung and Daniel Levy and Chak Ming Li and Rachel Lim and Molly Lin and Stephanie Lin and Mateusz Litwin and Theresa Lopez and Ryan Lowe and Patricia Lue and Anna Makanju and Kim Malfacini and Sam Manning and Todor Markov and Yaniv Markovski and Bianca Martin and Katie Mayer and Andrew Mayne and Bob McGrew and Scott Mayer McKinney and Christine McLeavey and Paul McMillan and Jake McNeil and David Medina and Aalok Mehta and Jacob Menick and Luke Metz and Andrey Mishchenko and Pamela Mishkin and Vinnie Monaco and Evan Morikawa and Daniel Mossing and Tong Mu and Mira Murati and Oleg Murk and David Mély and Ashvin Nair and Reiichiro Nakano and Rajeev Nayak and Arvind Neelakantan and Richard Ngo and Hyeonwoo Noh and Long Ouyang and Cullen O'Keefe and Jakub Pachocki and Alex Paino and Joe Palermo and Ashley Pantuliano and Giambattista Parascandolo and Joel Parish and Emy Parparita and Alex Passos and Mikhail Pavlov and Andrew Peng and Adam Perelman and Filipe de Avila Belbute Peres and Michael Petrov and Henrique Ponde de Oliveira Pinto and Michael and Pokorny and Michelle Pokrass and Vitchyr H. Pong and Tolly Powell and Alethea Power and Boris Power and Elizabeth Proehl and Raul Puri and Alec Radford and Jack Rae and Aditya Ramesh and Cameron Raymond and Francis Real and Kendra Rimbach and Carl Ross and Bob Rotsted and Henri Roussez and Nick Ryder and Mario Saltarelli and Ted Sanders and Shibani Santurkar and Girish Sastry and Heather Schmidt and David Schnurr and John Schulman and Daniel Selsam and Kyla Sheppard and Toki Sherbakov and Jessica Shieh and Sarah Shoker and Pranav Shyam and Szymon Sidor and Eric Sigler and Maddie Simens and Jordan Sitkin and Katarina Slama and Ian Sohl and Benjamin Sokolowsky and Yang Song and Natalie Staudacher and Felipe Petroski Such and Natalie Summers and Ilya Sutskever and Jie Tang and Nikolas Tezak and Madeleine B. Thompson and Phil Tillet and Amin Tootoonchian and Elizabeth Tseng and Preston Tuggle and Nick Turley and Jerry Tworek and Juan Felipe Cerón Uribe and Andrea Vallone and Arun Vijayvergiya and Chelsea Voss and Carroll Wainwright and Justin Jay Wang and Alvin Wang and Ben Wang and Jonathan Ward and Jason Wei and CJ Weinmann and Akila Welihinda and Peter Welinder and Jiayi Weng and Lilian Weng and Matt Wiethoff and Dave Willner and Clemens Winter and Samuel Wolrich and Hannah Wong and Lauren Workman and Sherwin Wu and Jeff Wu and Michael Wu and Kai Xiao and Tao Xu and Sarah Yoo and Kevin Yu and Qiming Yuan and Wojciech Zaremba and Rowan Zellers and Chong Zhang and Marvin Zhang and Shengjia Zhao and Tianhao Zheng and Juntang Zhuang and William Zhuk and Barret Zoph},
      year={2024},
      eprint={2303.08774},
      archivePrefix={arXiv},
      primaryClass={cs.CL}
}

@misc{laurençon2024matters,
      title={What matters when building vision-language models?}, 
      author={Hugo Laurençon and Léo Tronchon and Matthieu Cord and Victor Sanh},
      year={2024},
      eprint={2405.02246},
      archivePrefix={arXiv},
      primaryClass={cs.CV}
}

@inproceedings{picek2023snakeclef,
  title = {Overview of SnakeCLEF 2023: Snake Identification in Medically Important Scenarios},
  author = {Lukáš Picek and Rail Chamidullin and Marek Hrúz and Andrew M. Durso},
  booktitle = {CEUR Workshop Proceedings},
  volume = {3497},
  year = {2023},
  publisher = {CEUR-WS},
  url = {http://ceur-ws.org/Vol-3497/paper-168.pdf}
}

@inproceedings{sarto2022retrieval,
  title={Retrieval-augmented transformer for image captioning},
  author={Sarto, Sara and Cornia, Marcella and Baraldi, Lorenzo and Cucchiara, Rita},
  booktitle={Proceedings of the 19th international conference on content-based multimedia indexing},
  pages={1--7},
  year={2022}
}

@inproceedings{yasunaga2023retrieval,
  title={Retrieval-Augmented Multimodal Language Modeling},
  author={Yasunaga, Michihiro and Aghajanyan, Armen and Shi, Weijia and James, Richard and Leskovec, Jure and Liang, Percy and Lewis, Mike and Zettlemoyer, Luke and Yih, Wen-Tau},
  booktitle={International Conference on Machine Learning},
  pages={39755--39769},
  year={2023},
  organization={PMLR}
}

@inproceedings{chen2022murag,
  title={MuRAG: Multimodal Retrieval-Augmented Generator for Open Question Answering over Images and Text},
  author={Chen, Wenhu and Hu, Hexiang and Chen, Xi and Verga, Pat and Cohen, William},
  booktitle={Proceedings of the 2022 Conference on Empirical Methods in Natural Language Processing},
  pages={5558--5570},
  year={2022}
}

@inproceedings{guu2020retrieval,
  title={Retrieval augmented language model pre-training},
  author={Guu, Kelvin and Lee, Kenton and Tung, Zora and Pasupat, Panupong and Chang, Mingwei},
  booktitle={International conference on machine learning},
  pages={3929--3938},
  year={2020},
  organization={PMLR}
}

@misc{photutorial2021photos,
  author = {{Matic Broz}},
  title = {How many pictures are there (2024): Statistics, trends, and forecasts },
  year = {2024},
  howpublished = {\url{https://photutorial.com/photos-statistics/}},
  note = {Accessed: 2024-05-16}
}

@inproceedings{burns2022discovering,
  title={Discovering Latent Knowledge in Language Models Without Supervision},
  author={Burns, Collin and Ye, Haotian and Klein, Dan and Steinhardt, Jacob},
  booktitle={The Eleventh International Conference on Learning Representations},
  year={2022}
}

@article{yang2023dawn,
  title={The dawn of lmms: Preliminary explorations with gpt-4v (ision)},
  author={Yang, Zhengyuan and Li, Linjie and Lin, Kevin and Wang, Jianfeng and Lin, Chung-Ching and Liu, Zicheng and Wang, Lijuan},
  journal={arXiv preprint arXiv:2309.17421},
  volume={9},
  number={1},
  pages={1},
  year={2023}
}

@article{liu2024visual,
  title={Visual instruction tuning},
  author={Liu, Haotian and Li, Chunyuan and Wu, Qingyang and Lee, Yong Jae},
  journal={Advances in neural information processing systems},
  volume={36},
  year={2024}
}

@article{alayrac2022flamingo,
  title={Flamingo: a visual language model for few-shot learning},
  author={Alayrac, Jean-Baptiste and Donahue, Jeff and Luc, Pauline and Miech, Antoine and Barr, Iain and Hasson, Yana and Lenc, Karel and Mensch, Arthur and Millican, Katherine and Reynolds, Malcolm and others},
  journal={Advances in neural information processing systems},
  volume={35},
  pages={23716--23736},
  year={2022}
}

@article{caffagni2024wiki,
  title={Wiki-LLaVA: Hierarchical Retrieval-Augmented Generation for Multimodal LLMs},
  author={Caffagni, Davide and Cocchi, Federico and Moratelli, Nicholas and Sarto, Sara and Cornia, Marcella and Baraldi, Lorenzo and Cucchiara, Rita},
  journal={arXiv preprint arXiv:2404.15406},
  year={2024}
}

@article{deng2024mind2web,
  title={Mind2web: Towards a generalist agent for the web},
  author={Deng, Xiang and Gu, Yu and Zheng, Boyuan and Chen, Shijie and Stevens, Sam and Wang, Boshi and Sun, Huan and Su, Yu},
  journal={Advances in Neural Information Processing Systems},
  volume={36},
  year={2024}
}

@article{liu2023agentbench,
  title={Agentbench: Evaluating llms as agents},
  author={Liu, Xiao and Yu, Hao and Zhang, Hanchen and Xu, Yifan and Lei, Xuanyu and Lai, Hanyu and Gu, Yu and Ding, Hangliang and Men, Kaiwen and Yang, Kejuan and others},
  journal={arXiv preprint arXiv:2308.03688},
  year={2023}
}

@article{smith2022estimating,
  title={Estimating the number of diseases--the concept of rare, ultra-rare, and hyper-rare},
  author={Smith, CI Edvard and Bergman, Peter and Hagey, Daniel W},
  journal={Iscience},
  volume={25},
  number={8},
  year={2022},
  publisher={Elsevier}
}

@article{hu2024avis,
  title={Avis: Autonomous visual information seeking with large language model agent},
  author={Hu, Ziniu and Iscen, Ahmet and Sun, Chen and Chang, Kai-Wei and Sun, Yizhou and Ross, David and Schmid, Cordelia and Fathi, Alireza},
  journal={Advances in Neural Information Processing Systems},
  volume={36},
  year={2024}
}

@article{zheng2024judging,
  title={Judging llm-as-a-judge with mt-bench and chatbot arena},
  author={Zheng, Lianmin and Chiang, Wei-Lin and Sheng, Ying and Zhuang, Siyuan and Wu, Zhanghao and Zhuang, Yonghao and Lin, Zi and Li, Zhuohan and Li, Dacheng and Xing, Eric and others},
  journal={Advances in Neural Information Processing Systems},
  volume={36},
  year={2024}
}

@misc{compound-ai-blog,
  title={The Shift from Models to Compound AI Systems},
  author={Matei Zaharia and Omar Khattab and Lingjiao Chen and Jared Quincy Davis
          and Heather Miller and Chris Potts and James Zou and Michael Carbin
          and Jonathan Frankle and Naveen Rao and Ali Ghodsi},
  howpublished={\url{https://bair.berkeley.edu/blog/2024/02/18/compound-ai-systems/}},
  year={2024}
}

@article{li2023medical,
  title={A Comprehensive Study of GPT-4V's Multimodal Capabilities in Medical Imaging},
  author={Li, Yingshu and Liu, Yunyi and Wang, Zhanyu and Liang, Xinyu and Liu, Lingqiao and Wang, Lei and Cui, Leyang and Tu, Zhaopeng and Wang, Longyue and Zhou, Luping},
  journal={medRxiv},
  pages={2023--11},
  year={2023},
  publisher={Cold Spring Harbor Laboratory Press}
}

@article{jiang2024evaluating,
  title={Evaluating General Vision-Language Models for Clinical Medicine},
  author={Jiang, Yixing and Omiye, Jesutofunmi A and Zakka, Cyril and Moor, Michael and Gui, Haiwen and Alipour, Shayan and Mousavi, Seyed Shahabeddin and Chen, Jonathan H and Rajpurkar, Pranav and Daneshjou, Roxana},
  journal={medRxiv},
  pages={2024--04},
  year={2024},
  publisher={Cold Spring Harbor Laboratory Press}
}

@article{schmidgall2024agentclinic,
  title={AgentClinic: a multimodal agent benchmark to evaluate AI in simulated clinical environments},
  author={Schmidgall, Samuel and Ziaei, Rojin and Harris, Carl and Reis, Eduardo and Jopling, Jeffrey and Moor, Michael},
  journal={arXiv preprint arXiv:2405.07960},
  year={2024}
}


\appendix

\section*{Appendix }
\renewcommand{\thesection}{A.\arabic{section}}

\section{Additional experimental details}
\label{supp:additional-exp-details}
\subsection{Compute resources}
For the OpenAI models, the official API endpoints were called. One full run on the combination of INFOSEEK and SnakeCLEF uses around 3.5 Million tokens. This translates to a cost of approximately 20 to 110 USD, depending on the specific endpoint utilized (GPT-4o, GPT-4 Turbo, or GPT-4V). A single full run using GPT-4 Turbo to calculate the GPT-as-judge Accuracy involves processing approximately 3 million tokens, incurring a cost of around 10 USD per run. Experiments involving Idefics-2 were done on a single NVIDIA A100 GPU with 80G GPU memory, with the Idefics-2 model set to inference mode.

\subsection{Additional results on helpful and hurtful sets for RIR on INFOSEEK}
In Table~\ref{tab:net_gain_by_category}, we display the number of helpful and hurtful cases of RIR (i.e., cases where adding RIR switched a wrong answer into a right one, and vice versa). We observe the most helpful cases in categories that involves unique objects like facilities, buildings, locations, and plants~(e.g. a specific flower etc). By contrast, the most hurtful cases we find in categories that display everyday objects that are more common like foods and animals.

\begin{table}[bp]
\centering
\caption{RIR Helping and Hurting Cases on INFOSEEK with GPT-4o. In the \textbf{Helping} column lists the number of instances where the vanilla model provided incorrect answers, but the RIR-augmented model produced correct responses. Conversely, the \textbf{Hurting} column shows the number of cases where the RIR-augmented model gave incorrect answers that the vanilla model initially answered correctly. Higher values are highlighted with deeper colors to visualize trends. Positive net gain percentages are highlighted in \textbf{\textcolor[HTML]{0B8043}{green}}.}

\label{tab:net_gain_by_category}
\resizebox{0.75\textwidth}{!}{%
\begin{tabular}{ccccc}
\toprule
\textbf{Category} & \textbf{Helping} & \textbf{Hurting} & \textbf{Net Gain} & \textbf{Net Gain (\%)} \\ \midrule
facility & \cellcolor[HTML]{76C8A0}41 & \cellcolor[HTML]{FCEFEE}5 & \cellcolor[HTML]{FFD666}36 & {\color[HTML]{0B8043} \textbf{24.00}} \\
building & \cellcolor[HTML]{6FC59B}43 & \cellcolor[HTML]{FBE9E7}7 & \cellcolor[HTML]{FFD666}36 & {\color[HTML]{0B8043} \textbf{24.00}} \\
location & \cellcolor[HTML]{A5DBC0}27 & \cellcolor[HTML]{FEF6F5}3 & \cellcolor[HTML]{FFE18E}24 & {\color[HTML]{0B8043} \textbf{16.00}} \\
plant & \cellcolor[HTML]{C6E8D8}17 & \cellcolor[HTML]{FCEFEE}5 & \cellcolor[HTML]{FFEBB5}12 & {\color[HTML]{0B8043} \textbf{8.00}} \\
others & \cellcolor[HTML]{C0E6D3}19 & \cellcolor[HTML]{FAE2E0}9 & \cellcolor[HTML]{FFEDBB}10 & {\color[HTML]{0B8043} \textbf{6.67}} \\
sport & \cellcolor[HTML]{D0ECDF}14 & \cellcolor[HTML]{FBE9E7}7 & \cellcolor[HTML]{FFF0C5}7 & {\color[HTML]{0B8043} \textbf{4.67}} \\
vehicle & \cellcolor[HTML]{C6E8D8}17 & \cellcolor[HTML]{F8D8D5}12 & \cellcolor[HTML]{FFF2CB}5 & {\color[HTML]{0B8043} \textbf{3.33}} \\
organization and company & \cellcolor[HTML]{E5F5ED}8 & \cellcolor[HTML]{FEF6F5}3 & \cellcolor[HTML]{FFF2CB}5 & {\color[HTML]{0B8043} \textbf{3.33}} \\
objects & \cellcolor[HTML]{D4EEE1}13 & \cellcolor[HTML]{F9DFDC}10 & \cellcolor[HTML]{FFF3D2}3 & {\color[HTML]{0B8043} \textbf{2.00}} \\
animal & \cellcolor[HTML]{CDEBDC}15 & \cellcolor[HTML]{F6CECB}15 & \cellcolor[HTML]{FFF6DC}0 & 0.00 \\
food & \cellcolor[HTML]{E8F6EF}7 & \cellcolor[HTML]{F4C5C0}18 & \cellcolor[HTML]{FFFFFF}-11 & -7.33 \\ \bottomrule
\end{tabular}%
}
\end{table}

\subsection{Human evaluation details}
\label{supp:human-eval}

\paragraph{INFOSEEK} We randomly sampled $9$ data samples from each of the $11$ categories in INFOSEEK, generating $198$ responses from GPT-4o and GPT-4o$_{RIR}$. These responses, along with their GPT-as-judge judgments, were manually evaluated for 1) whether the GPT-as-judge Accuracy metric is reliable, \textit{i.e.}, how well does the GPT-as-judge judgment align with human judgment, and 2) whether RIR is directly providing the answer to the question. Our evaluation revealed that GPT-as-judge is accurate in evaluating the correctness of model-generated answers, with only $6$ out of $198$ evaluations ($3.03\%$) diverging from human judgment. In addition, We discovered that in $99\%$ of the cases we investigated, RIR did not provide a direct final answer, but offered additional contexts for answering the question, implying that on INFOSEEK RIR mainly helps by stimulating MLLMs' own textual world knowledge.

\paragraph{SnakeCLEF} From our SnakeCLEF dataset, we randomly sampled $50$ data samples and collected the responses from GPT-4o$_{RIR}$. We manually evaluate the data samples to determine the proportion of questions for which RIR provides direct answers. We found that out of $50$, in $9$ samples ($18\%$) the RIR screenshot contains the correct genus of the snake, and in another $8$ sample ($16\%$) it contains the correct common name of the snake (\textit{e.g.} when ground truth is ``virginia valeriae'' the RIR screenshot contained ``Smooth Earth Snake''). These results show that RIR mainly helps by providing visual knowledge on SnakeCLEF.

\subsection{Prompt details for RIR}\label{supp:prompt-rir}
RIR returns reverse image search results which we capture in a screenshot. The screenshot image is provided as context to the down-stream MLLM together with a layout explanation prompt as illustrated below. 

\begin{quote}
\texttt{In the screenshot, the large image on the left is the query image for a reverse image search.
The smaller images on the right and their titles are the top hits from the search.}
\end{quote}

Overall, the full RIR pipeline consists of the following steps: 
\begin{enumerate}
    \item Reverse image search of query image 
    \item Screenshot capture of results page
    \item Message composition to send to down-stream MLLM, featuring:
    \begin{itemize}
        \item System prompt
        \item Screenshot image
        \item Layout explanation
        \item Query image
        \item Query text
    \end{itemize}
    \item Return response from MLLM
\end{enumerate}
\subsection{Prompt details for analysis with oracle-provided entities}\label{supp:prompt-oracle}

As sketched in Figure~\ref{fig:knowledge}, we discovered that RIR's benefit in knowledge-intensive VQA primarily comes from helping the MLLM accessing its own world knowledge, rather than just by supplying external knowledge that is sufficient to answer the visual question.
To support this claim, we provide an additional analysis where for the INFOSEEK benchmark, we convert visual questions $(v,q)$ to rephrased text-only questions $q'$ that already contain the entity of interest. When converting, we use the following template: 

\begin{quote}
\texttt{Imagine that you are presented with an image of \{\{entity\}\}. Answer the following question: \{\{question\}\}}. 
\end{quote}

Here, \texttt{\{\{entity\}\}} and \texttt{\{\{question\}\}} are replaced with the ground truth entity name and the original query, respectively. For example, the question shown in Figure~\ref{fig:knowledge} will be converted into 
\begin{quote}
\texttt{Imagine that you are presented with an image of Bouzov Castle. Answer the following question: What country does this building belong to?}
\end{quote}

The ground truth entity name is collected using the ground truth entity Wikidata ID of each question, provided by the INFOSEEK dataset. We convert the Wikidata IDs to their entity names by querying Wikidata and scraping the title of the page associated with each Wikidata ID.

\subsection{Does a visual search agent outperform RIR?}
\label{supp:agent}
Here, we consider in which cases a basic, MLLM-powered agent is able to improve upon the RIR-augmented performance.
While many types of agents are conceivable, the main capability we care about here is the decision to call RIR, or not.
For $n$ samples of a given task (e.g., a VQA problem), we distinguish three cases: There are $p$ \emph{helpful} cases where calling RIR is counterfactually flipping a false prediction to a correct prediction. There are $q$ \emph{hurtful} cases, where calling RIR is flipping otherwise correct predictions into wrong ones. Finally, there are $n - q - p$ neutral cases, where the decision to call RIR does not affect the prediction correctness. Also, let $a, b \in [0,1]$ denote the fraction of helpful and hurtful cases that the agent misclassifies, i.e., $a$ is the fraction of helpful cases where RIR was not called, and $b$ is the fraction of hurtful cases where RIR was called.
In Figure~\ref{fig:agent-help}, we illustrate how the decision to call RIR is challenged by the imbalance of $p$ and $q$. If $p$ is very large, then it is hard to overcome the inequality shown in Figure~\ref{fig:agent-help}. For example, in the case that $\frac{p}{q} = 5$, i.e., there are five times more helpful than hurtful cases, and $a = b = 0.2$, i.e., the agent only correctly identifies 80\% of the hurtful and helpful cases (and acts accordingly), then $\frac{p}{q} > \frac{1-b}{a}$, hence the agent would be sub-par to the approach that always defaults to call RIR.

Next, we present result for a simple GPT-4o-based agent that can decide to either call RIR or not, and that given search results from a call runs a consistency check that the search results are relevant before giving the final answer. Table~\ref{tab:agent-table} compares the performance of such an agent (row ``Decide + Consistency Check'') against the approach that defaults to RIR (row ``Always use RIR''). In addition, we show ablations of the agent using only one of the two features. The results suggest that the default RIR variant is not outperformed by these agent variants in terms of accuracy, which can be explained by those agents high error rates in detecting helpful and hurtful cases. 

\begin{table}[h!]
\centering
\caption{Comparison of a GPT-4o-based agent that can call RIR (Decide + Consistency Check) against the approach to default to always using RIR. Additional ablations are shown for an agent that only decides to use RIR, and one that only consistency checks the search results. Due to too high error rates in identifying helpful and hurtful cases of RIR, the agent does not outperform the default RIR approach.}
\label{tab:agent-table}
\begin{tabular}{lccc}
\toprule
\textbf{Configuration} & \textbf{\begin{tabular}[c]{@{}c@{}}Error Rate \\ Helping ($a$)\end{tabular}} & \textbf{\begin{tabular}[c]{@{}c@{}}Error Rate \\ Hurting ($b$)\end{tabular}} & \textbf{\begin{tabular}[c]{@{}c@{}}Overall Accuracy\\ GPT-as-judge\end{tabular}} \\ \midrule
Always use RIR & 0.00\% & 100.00\% & \textbf{46.91}\% \\
Decide + Consistency Check & 57.92\% & 28.72\% & 43.21\% \\
Decide only & 17.19\% & 67.02\% & 46.48\% \\
Consistency Check only & 53.85\% & 39.36\% & 43.15\% \\ \bottomrule
\end{tabular}
\end{table}

\subsection{Extended Results with Confidence Intervals}
\label{sec:bootstrap}
We employed a bootstrapping approach to estimate the variability of our model performance metrics. For INFOSEEK, we generated 1000 bootstrap samples for each category, where each bootstrap sample was created by randomly drawing 150 (size of each category) samples with replacements from the original category data. For SnakeCLEF, we generated 1000 bootstrap samples for the full 300 data samples, where each bootstrap sample was created by randomly drawing 300 samples with replacements. 
We then evaluated the models' performance metrics on each bootstrap sample and calculated the 95\% confidence intervals for these metrics. The results are presented in Tables~\ref{tab:bootstrap_infoseek},~\ref{tab:bootstrap_infoseek_2}~and~\ref{tab:bootstrap_snake}. 

\begin{table}[h]
\centering
\caption{Experiment results on INFOSEEK. The models and evaluation metrics used are described in Section~\ref{sec:backbone_models} and Section~\ref{sec:datasets}. We report the 95\% confidence intervals, derived from 1000 bootstrap samples across all eleven categories.}
\label{tab:bootstrap_infoseek}
\resizebox{0.8\textwidth}{!}{%
\begin{tabular}{@{}lllllll@{}}
\toprule
\multicolumn{1}{l|}{\textbf{Model}} & \multicolumn{1}{c|}{\textbf{Avg.}} & \multicolumn{1}{c}{\textbf{Building}} & \multicolumn{1}{c}{\textbf{Animal}} & \multicolumn{1}{c}{\textbf{Plant}} & \multicolumn{1}{c}{\textbf{Location}} & \multicolumn{1}{c}{\textbf{Food}} \\ \midrule
\multicolumn{7}{c}{\textit{GPT-as-judge Accuracy (\%)}} \\ \midrule
\multicolumn{1}{l|}{Idefics2} & \multicolumn{1}{l|}{$(15.70,~19.09)$} & $(5.33,~14.00)$ & $(0.67,~6.67)$ & $(0.67,~5.33)$ & $(7.33,~18.00)$ & $(25.33,~39.33)$ \\
\multicolumn{1}{l|}{GPT4V} & \multicolumn{1}{l|}{$(29.09,~33.58)$} & $(12.67,~26.00)$ & $(23.33,~38.00)$ & $(4.00,~12.67)$ & $(18.00,~31.33)$ & $(33.33,~50.00)$ \\
\multicolumn{1}{l|}{GPT-4-turbo} & \multicolumn{1}{l|}{$(34.24,~38.73)$} & $(22.67,~37.33)$ & $(30.00,~45.33)$ & $(7.33,~17.33)$ & $(28.00,~42.67)$ & $(40.00,~55.33)$ \\
\multicolumn{1}{l|}{GPT-4o} & \multicolumn{1}{l|}{$(36.73,~41.33)$} & $(28.00,~42.67)$ & $(26.00,~41.33)$ & $(7.33,~18.00)$ & $(28.67,~43.35)$ & $(46.67,~62.67)$ \\
\multicolumn{1}{l|}{Idefics2$_{RIR}$} & \multicolumn{1}{l|}{$(16.85,~20.67)$} & $(13.33,~26.00)$ & $(2.00,~8.67)$ & $(0.67,~5.33)$ & $(11.33,~23.35)$ & $(28.67,~42.67)$ \\
\multicolumn{1}{l|}{GPT4V$_{RIR}$} & \multicolumn{1}{l|}{$(42.18,~47.03)$} & $(46.00,~62.00)$ & $(28.00,~42.68)$ & $(14.00,~26.02)$ & $(38.67,~54.67)$ & $(37.33,~54.00)$ \\
\multicolumn{1}{l|}{GPT-4-turbo$_{RIR}$} & \multicolumn{1}{l|}{$(44.00,~48.79)$} & $(51.33,~67.33)$ & $(31.33,~46.67)$ & $(14.00,~26.00)$ & $(44.00,~60.00)$ & $(36.00,~51.33)$ \\
\multicolumn{1}{l|}{GPT-4o$_{RIR}$} & \multicolumn{1}{l|}{$(44.61,~49.33)$} & $(51.33,~66.00)$ & $(25.98,~41.33)$ & $(14.67,~26.67)$ & $(43.33,~60.00)$ & $(39.32,~56.00)$ \\ \midrule
\multicolumn{7}{c}{\textit{Answer-in-prediction Recall (\%)}} \\ \midrule
\multicolumn{1}{l|}{Idefics2} & \multicolumn{1}{l|}{$(12.60,~15.88)$} & $(2.67,~10.00)$ & $(1.33,~8.67)$ & $(1.33,~8.67)$ & $(6.00,~16.02)$ & $(14.00,~28.00)$ \\
\multicolumn{1}{l|}{GPT4V} & \multicolumn{1}{l|}{$(27.39,~31.88)$} & $(12.00,~24.68)$ & $(20.00,~34.00)$ & $(12.00,~24.00)$ & $(18.00,~32.00)$ & $(22.67,~36.67)$ \\
\multicolumn{1}{l|}{GPT-4-turbo} & \multicolumn{1}{l|}{$(30.91,~35.15)$} & $(22.67,~37.33)$ & $(22.67,~36.67)$ & $(14.00,~27.33)$ & $(27.33,~42.00)$ & $(28.00,~44.00)$ \\
\multicolumn{1}{l|}{GPT-4o} & \multicolumn{1}{l|}{$(33.64,~38.18)$} & $(26.00,~40.67)$ & $(22.67,~38.00)$ & $(21.33,~36.00)$ & $(28.00,~43.33)$ & $(31.33,~47.33)$ \\
\multicolumn{1}{l|}{Idefics2$_{RIR}$} & \multicolumn{1}{l|}{$(13.94,~17.33)$} & $(8.67,~19.35)$ & $(2.67,~10.00)$ & $(1.33,~8.67)$ & $(9.33,~20.00)$ & $(16.67,~31.33)$ \\
\multicolumn{1}{l|}{GPT4V$_{RIR}$} & \multicolumn{1}{l|}{$(38.30,~42.97)$} & $(39.33,~55.33)$ & $(22.67,~37.33)$ & $(28.67,~44.67)$ & $(34.00,~49.33)$ & $(29.98,~44.67)$ \\
\multicolumn{1}{l|}{GPT-4-turbo$_{RIR}$} & \multicolumn{1}{l|}{$(38.79,~43.52)$} & $(44.67,~60.67)$ & $(25.32,~39.33)$ & $(29.33,~44.02)$ & $(40.00,~56.00)$ & $(26.67,~42.67)$ \\
\multicolumn{1}{l|}{GPT-4o$_{RIR}$} & \multicolumn{1}{l|}{$(40.36,~45.09)$} & $(49.33,~64.67)$ & $(22.00,~36.67)$ & $(30.00,~46.00)$ & $(37.33,~52.67)$ & $(30.67,~44.67)$ \\ \bottomrule
\end{tabular}%
}
\end{table}

\begin{table}[h]
\centering
\caption{Experiment results on INFOSEEK (Continued). The models and evaluation metrics used are described in Section~\ref{sec:backbone_models} and Section~\ref{sec:datasets}. We report the 95\% confidence intervals, derived from 1000 bootstrap samples across all eleven categories.}
\label{tab:bootstrap_infoseek_2}
\resizebox{0.8\textwidth}{!}{%
\begin{tabular}{lllllll}
\toprule
\multicolumn{1}{l|}{\textbf{Model}} & \multicolumn{1}{c}{\textbf{OC}} & \multicolumn{1}{c}{\textbf{Facility}} & \multicolumn{1}{c}{\textbf{Vehicle}} & \multicolumn{1}{c}{\textbf{Objects}} & \multicolumn{1}{c}{\textbf{Sport}} & \multicolumn{1}{c}{\textbf{Others}} \\ \midrule
\multicolumn{7}{c}{\textit{GPT-as-judge Accuracy (\%)}} \\ \midrule
\multicolumn{1}{l|}{Idefics2} & $(4.00,~12.67)$ & $(4.67,~14.00)$ & $(36.67,~52.67)$ & $(5.33,~15.33)$ & $(40.65,~56.00)$ & $(5.98,~16.00)$ \\
\multicolumn{1}{l|}{GPT4V} & $(12.00,~24.67)$ & $(19.33,~33.33)$ & $(55.33,~71.33)$ & $(16.67,~30.00)$ & $(50.00,~66.00)$ & $(25.33,~40.00)$ \\
\multicolumn{1}{l|}{GPT-4-turbo} & $(16.67,~30.67)$ & $(32.67,~47.35)$ & $(52.67,~68.67)$ & $(12.00,~24.67)$ & $(52.67,~67.33)$ & $(30.67,~47.33)$ \\
\multicolumn{1}{l|}{GPT-4o} & $(16.67,~30.00)$ & $(32.67,~48.67)$ & $(58.67,~72.67)$ & $(15.33,~28.00)$ & $(55.33,~71.33)$ & $(38.65,~53.35)$ \\
\multicolumn{1}{l|}{Idefics2$_{RIR}$} & $(2.00,~9.33)$ & $(8.00,~18.67)$ & $(36.00,~52.00)$ & $(1.33,~7.33)$ & $(38.67,~55.33)$ & $(7.33,~17.33)$ \\
\multicolumn{1}{l|}{GPT4V$_{RIR}$} & $(22.00,~36.67)$ & $(46.67,~62.67)$ & $(55.33,~71.33)$ & $(14.67,~28.00)$ & $(56.00,~71.33)$ & $(50.00,~65.33)$ \\
\multicolumn{1}{l|}{GPT-4-turbo$_{RIR}$} & $(19.33,~34.00)$ & $(54.00,~70.00)$ & $(57.98,~73.33)$ & $(14.00,~26.67)$ & $(58.00,~74.00)$ & $(48.67,~64.00)$ \\
\multicolumn{1}{l|}{GPT-4o$_{RIR}$} & $(19.33,~32.67)$ & $(56.67,~72.00)$ & $(61.33,~76.00)$ & $(16.67,~30.00)$ & $(61.33,~74.68)$ & $(44.00,~60.67)$ \\ \midrule
\multicolumn{7}{c}{\textit{Answer-in-prediction Recall (\%)}} \\ \midrule
\multicolumn{1}{l|}{Idefics2} & $(4.00,~12.67)$ & $(4.67,~14.00)$ & $(18.67,~34.00)$ & $(4.00,~13.33)$ & $(40.67,~56.67)$ & $(5.33,~14.00)$ \\
\multicolumn{1}{l|}{GPT4V} & $(19.33,~33.33)$ & $(19.33,~34.00)$ & $(30.00,~46.02)$ & $(16.67,~30.00)$ & $(52.67,~68.67)$ & $(26.00,~41.35)$ \\
\multicolumn{1}{l|}{GPT-4-turbo} & $(19.33,~34.00)$ & $(31.33,~46.67)$ & $(24.00,~38.67)$ & $(14.65,~26.68)$ & $(54.00,~69.33)$ & $(24.67,~40.00)$ \\
\multicolumn{1}{l|}{GPT-4o} & $(20.00,~34.67)$ & $(29.33,~44.67)$ & $(29.33,~44.68)$ & $(17.33,~30.67)$ & $(57.33,~73.33)$ & $(31.33,~47.33)$ \\
\multicolumn{1}{l|}{Idefics2$_{RIR}$} & $(0.00,~3.33)$ & $(8.00,~18.67)$ & $(19.98,~34.00)$ & $(2.67,~10.00)$ & $(44.00,~60.00)$ & $(4.67,~14.00)$ \\
\multicolumn{1}{l|}{GPT4V$_{RIR}$} & $(23.33,~37.35)$ & $(39.33,~55.33)$ & $(32.00,~47.33)$ & $(18.00,~32.00)$ & $(58.00,~72.67)$ & $(40.00,~55.33)$ \\
\multicolumn{1}{l|}{GPT-4-turbo$_{RIR}$} & $(20.67,~34.67)$ & $(40.00,~56.67)$ & $(32.00,~47.33)$ & $(13.33,~26.67)$ & $(59.33,~74.67)$ & $(38.00,~53.33)$ \\
\multicolumn{1}{l|}{GPT-4o$_{RIR}$} & $(25.33,~40.67)$ & $(41.33,~58.00)$ & $(30.00,~46.00)$ & $(16.67,~30.67)$ & $(62.67,~76.67)$ & $(42.00,~58.00)$ \\ \bottomrule
\end{tabular}%
}
\end{table}

\begin{table}[h]
\centering
\caption{Experiment results on SnakeCLEF. The models and evaluation metrics used are described in Section~\ref{sec:backbone_models} and Section~\ref{sec:datasets}. We report the 95\% confidence intervals, derived from 1000 bootstrap samples.}
\label{tab:bootstrap_snake}
\resizebox{0.8\textwidth}{!}{%
\begin{tabular}{lcccc}
\toprule
\textbf{Model} & \textbf{Binomial-EM} & \textbf{Genus-EM} & \textbf{Binomial-Recall} & \textbf{Genus-Recall} \\ \midrule
Idefics2 & $(0.00,~0.00)$ & $(0.00,~1.67)$ & $(0.00,~1.00)$ & $(2.33,~7.33)$ \\
GPT4V & $(0.33,~3.33)$ & $(2.67,~7.67)$ & $(0.67,~3.67)$ & $(3.67,~9.00)$ \\
GPT-4-turbo & $(1.00,~4.67)$ & $(6.00,~12.01)$ & $(1.33,~5.00)$ & $(6.67,~13.67)$ \\
GPT4o & $(2.67,~8.01)$ & $(15.67,~25.00)$ & $(3.00,~8.01)$ & $(16.00,~25.00)$ \\ \midrule
Idefics2$_{RIR}$ & $(0.00,~0.00)$ & $(0.33,~2.67)$ & $(0.67,~3.67)$ & $(7.33,~14.33)$ \\
GPT4V$_{RIR}$ & $(7.67,~14.67)$ & $(19.33,~28.67)$ & $(8.00,~15.33)$ & $(19.33,~29.33)$ \\
GPT-4-turbo$_{RIR}$ & $(8.00,~15.33)$ & $(20.33,~29.67)$ & $(8.66,~16.00)$ & $(20.33,~30.00)$ \\
GPT4o$_{RIR}$ & $(9.00,~16.00)$ & $(20.67,~30.67)$ & $(9.33,~16.67)$ & $(21.33,~31.00)$ \\ \bottomrule
\end{tabular}%
}
\end{table}

\subsection{Ablation Study on RIR.}
\label{sec:ablation}

To understand what component of RIR augmentation is contributing to the performance gain, we perform an ablation where either the returned images or the returned text (titles and captions of images) are masked. We ran this ablation on a random but category-stratified subset of the INFOSEEK dataset of 550 samples. The results are shown in Table~\ref{tab:rir-ablation}. We find that both the images and text of the RIR result are providing signal to the MLLM (here GPT-4o) to improve upon the baseline MLLM. While we observe no clear effect from masking the image on this data subset, masking the text reduces the performance gain of RIR to 60\% of the gain that full RIR achieves.

\begin{table}[h]
\centering
\caption{Ablation of RIR components. We either mask the images or mask the text component of the RIR search result capture. Notably, GPT-4o is capable of leveraging both parts of the RIR augmentation to improve the baseline performance.}
\label{tab:rir-ablation}
\resizebox{0.3\textwidth}{!}{%
\begin{tabular}{lc}
\toprule
\textbf{Method} & \textbf{Acc} \\ \midrule
GPT-4o & 42.55 \\
GPT-4o RIR & 48.00 \\
GPT-4o RIR mask image & 48.00 \\
GPT-4o RIR mask text & 45.82 \\ \bottomrule
\end{tabular}%
}
\end{table}

\end{document}